\documentclass[letterpaper]{article} 
\usepackage{aaai24}  
\usepackage{times}  
\usepackage{helvet}  
\usepackage{courier}  
\usepackage[hyphens]{url}  
\usepackage{graphicx} 
\usepackage{natbib}  
\usepackage{caption} 
\usepackage{algorithm}
\usepackage{algorithmic}

\usepackage{newfloat}
\usepackage{listings}
\DeclareCaptionStyle{ruled}{labelfont=normalfont,labelsep=colon,strut=off} 
\floatstyle{ruled}
\newfloat{listing}{tb}{lst}{}
\floatname{listing}{Listing}
\usepackage{multirow}
\newtheorem{hyp}{Hypothesis}
\usepackage{amsmath}
\usepackage{hyperref}
\title{Jointly Exploring Client Drift and Catastrophic Forgetting in Dynamic Learning}
\author {
    Niklas Babendererde\textsuperscript{\rm 1},
    Moritz Fuchs\textsuperscript{\rm 1},
    Camila Gonzalez\textsuperscript{\rm 1},
    Yuri Tolkach\textsuperscript{\rm 2},
    Anirban Mukhopadhyay\textsuperscript{\rm 1}
}
\affiliations {
    \textsuperscript{\rm 1}Technical University Darmstadt\\
    \textsuperscript{\rm 2} University of Cologne\\
}

\usepackage{bibentry}

\begin{document}

\maketitle

\begin{abstract}
Federated and Continual Learning have emerged as potential paradigms for the robust and privacy-aware use of Deep Learning in dynamic environments. However, Client Drift and Catastrophic Forgetting are fundamental obstacles to guaranteeing consistent performance. Existing work only addresses these problems separately, which neglects the fact that the root cause behind both forms of performance deterioration is connected. We propose a unified analysis framework for building a controlled test environment for Client Drift -- by perturbing a defined ratio of clients -- and Catastrophic Forgetting -- by shifting all clients with a particular strength. Our framework further leverages this new combined analysis by generating a 3D landscape of the combined performance impact from both. We demonstrate that the performance drop through Client Drift, caused by a certain share of shifted clients, is correlated to the drop from Catastrophic Forgetting resulting from a corresponding shift strength. Correlation tests between both problems for Computer Vision (CelebA) and Medical Imaging (PESO) support this new perspective, with an average Pearson rank correlation coefficient of over 0.94.
Our framework's novel ability of combined spatio-temporal shift analysis allows us to investigate how both forms of distribution shift behave in mixed scenarios, opening a new pathway for better generalization. We show that a combination of moderate Client Drift and Catastrophic Forgetting can even improve the performance of the resulting model (causing a "Generalization Bump") compared to when only one of the shifts occurs individually.
We apply a simple and commonly used method from Continual Learning in the federated setting and observe this phenomenon to be reoccurring, leveraging the ability of our framework to analyze existing and novel methods for Federated and Continual Learning.
\end{abstract}

\section{Introduction}
Learning system deployment is often hindered by robustness issues stemming from \emph{static world assumptions}. While helpful for evaluation simplicity, these assumptions don't apply to our dynamic reality, where data distributions shift due to spatio-temporal \emph{covariate and population shifts}~\cite{jaeger2022call}.
In privacy-sensitive applications such as AI-assisted healthcare or camera-based object detection in autonomous driving, this problem becomes even more pronounced as regulations such as the \emph{HIPAA}\footnote{https://www.hhs.gov/hipaa/index.html} (in the USA) or \emph{GDPR}\footnote{https://gdpr.eu/} (in Europe) limit the availability of data. This typically means that samples acquired at different hospitals or vehicles cannot be stored centrally and are only available for a certain period. It is, therefore, challenging to cover all potential input variations, as would be required for a reliable and safety-aware application.

\begin{figure}[h]
    \centering
    \includegraphics[width=0.48\textwidth]{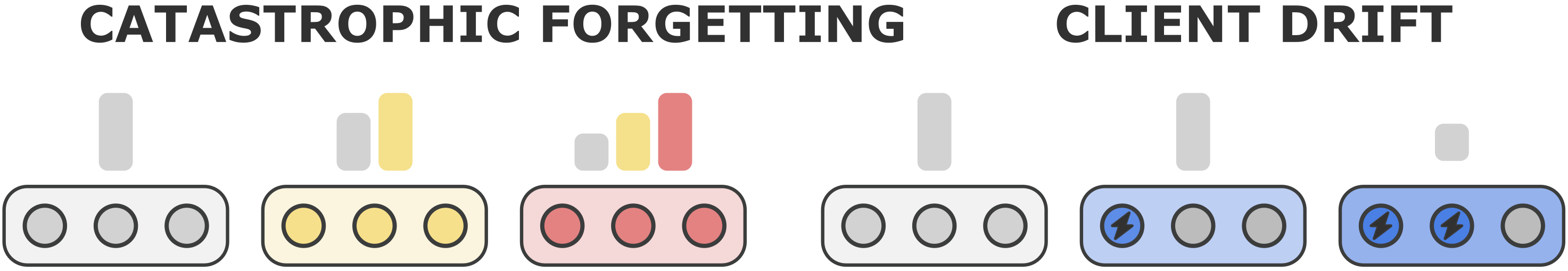}
    \caption{\textbf{Catastrophic Forgetting}: The distribution of the training data (dots in the box) changes over time (horizontal axis). This causes the performance (bar plots) on the old distribution to decrease. \textbf{Client Drift}: The data shifts spatially as a portion of the clients acquires data with certain characteristics, decreasing the performance of the federated model.}
    \label{fig:overview}
\end{figure}

Depending on the data availability constraints, two solutions have been proposed: \emph{Federated Learning \textbf{FL}} updates models in a decentralized fashion, allowing learning from spatially dynamic cases that are stored locally at the individual clients. \emph{Continual Learning \textbf{(CL)}} instead addresses the temporal dimension by training models sequentially over time.
Unfortunately, spatial- as well as temporal distribution shifts strongly degrade the performance of the resulting models, as Figure~\ref{fig:overview} visualizes.

Many factors can result in distribution shift in different application domains. For instance, when training a federated model for medical imaging, the presence of acquisition artifacts, like MRI-ghosting caused by a malfunctioning scanner, can degrade the performance of the central model~\cite{dalmics2017using}. In the context of camera-based object detection for autonomous driving, issues such as defective or dirty cameras can cause data shifts in specific vehicles, leading to a decline in the central model's effectiveness. This is known as \emph{Client Drift \textbf{(CD)}}~\cite{8}.

Likewise, a model trained to identify pulmonary lesions in chest CT scans might experience failures as the disease phenotype changes -- such as seen with the emergence of Covid-19~\cite{gonzalez2022distance}. After an update, the model may adapt to detecting Covid-19 cases but lose its ability to identify lesions caused by bacterial pneumonia. In autonomous driving, temporal weather fluctuations can erode object detection performance over time, posing risks to human lives. This phenomenon is termed \emph{Catastrophic Forgetting \textbf{(CF)}}~\cite{catastrophic_forgetting}.

\begin{figure*}[ht]
    \centering
    \includegraphics[width=0.9\textwidth]{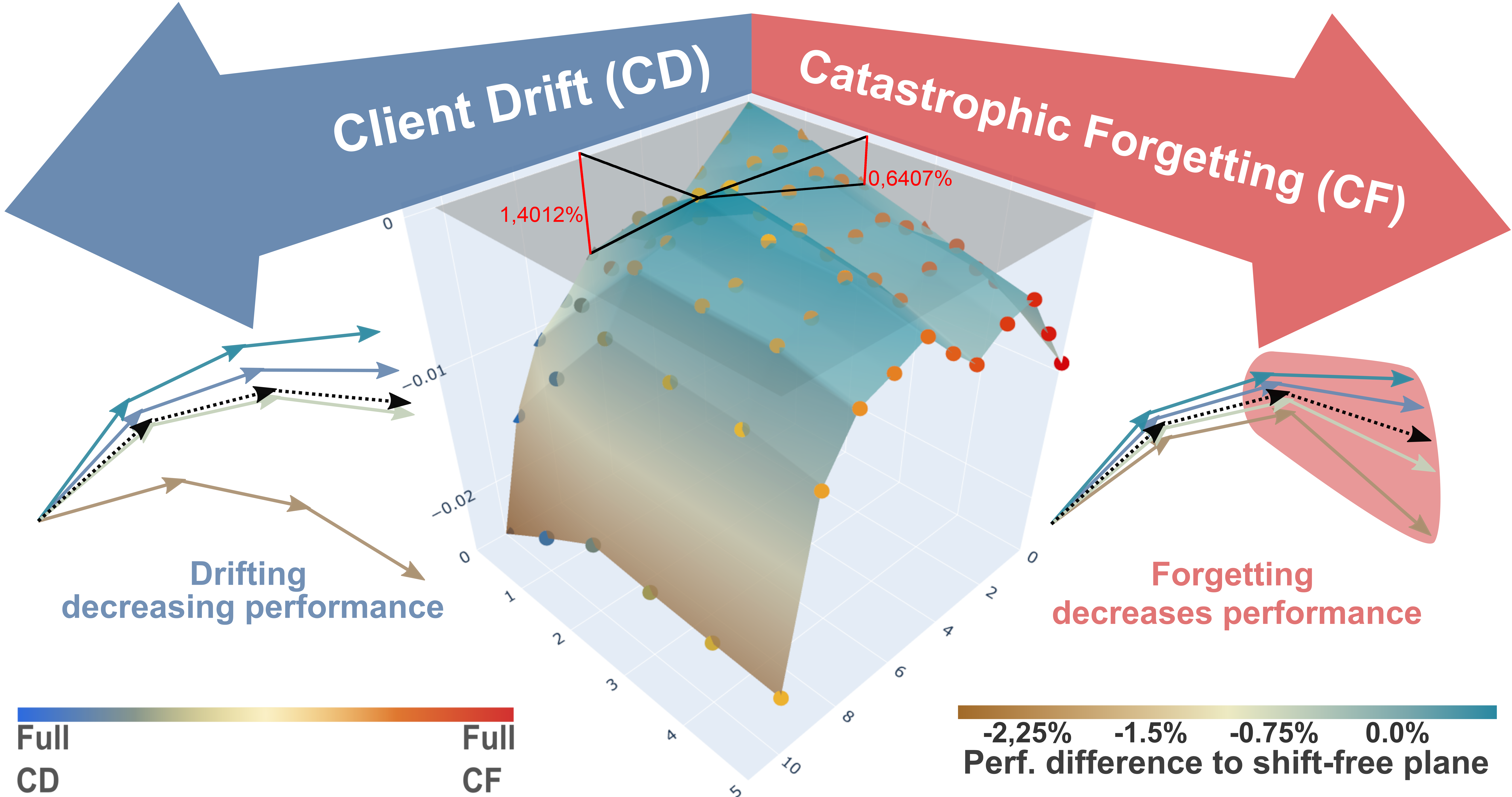}
    \caption{Combined visualization of the performance drop from CD and CF in mixed scenarios. The dimensions in the ground plane represent the ratio of shifted clients in CD (blue) and the shift in CF (red). The vertical dimension indicates the resulting performance difference on the classification dataset CelebA, which is measured in accuracy. While the values on the horizontal axes of the figure represent pure CD- or CF scenarios, the middle visualizes the outcome of mixed scenarios. (Left) Visualization of CD during training. (Right) Illustration of CF occurring in the red part of the training period. The black lines originate from the peak of the \emph{"generalization bump" from increased generalization,} which shows as improved performance when moderate CD and CF are combined. The red lines and numbers indicate the relative performance improvements to individual CD/CF.}
    \label{fig:mesh1}
\end{figure*}
For the safe and privacy-aware deployment of Deep Learning models in real-world applications, \emph{a holistic strategy must address both of these concerns}, as they typically occur concurrently.
However, in the current research landscape, these problems are studied separately -- as shown in Table \ref{tab:relatedwork} in the supplementary material -- even though all approaches try to solve the same underlying problem of distribution shift.

Several strategies have been studied for addressing CD, such as \emph{applying penalties on outliers}~\cite{12} or \emph{adding momentum to the server updates}~\cite{9}. Similarly, \emph{rehearsal methods} avoid CF by \emph{storing a portion of previous samples and interleaving them in later training}~\cite{rebuffi2017icarl}. This interleaving ensures that the model is concurrently learning with old and new cases, as both are present during training. Other work applies \emph{parameter tuning}~\cite{25, 26}. These methods are specifically designed for the challenge of CF through temporal dynamics and do not consider that data may be stored in different locations.

Addressing these problems separately, as in existing work, is a missed opportunity and disregards the fact that in real dynamic environments, distribution shifts occur spatially as well as temporally and interact with each other. FL is the preferred form of training models in privacy-protected settings, and updating the model as time goes on is unavoidable. Consequently, \emph{a unified understanding of spatial- and temporal dynamics is required}.

To understand the relationship between spatial and temporal distribution shifts, we propose \textbf{a unified framework that emulates combined CD and CF in a controlled fashion}. Our framework allows us to simulate a quantifiable spatial or temporal shift from controlled corruptions in the data and analyze the impact with respect to model deterioration. In the CD scenario, we shift a share of clients affected by the distribution shift, represented as a covariate shift.
Similarly, we emulate CF to shifting the distribution for all clients by a certain factor over time.
We apply our analysis framework to a common computer vision classification task (CelebA by~\citet{liu2015faceattributes}) and a medical segmentation use case (PESO by~\citet{bulten2019epithelium}). The results show that the performance drop through CD is correlated to the drop from CF, as we found the \textbf{average Pearson rank correlation coefficient between both problems is over 0.94}.

Moreover, to analyze the interaction of CD and CF, our framework allows controlled, joint spatial and temporal shifts in the same experiment, and we leverage this to combine different strengths of CD and CF and plot the resulting landscape of performance impact in Figure~\ref{fig:mesh1}. The illustration shows an improvement in scenarios of mixed and moderate CD and CF. We refer to this phenomenon as a "Generalization Bump".

In addition to this, we show that our analysis framework can also be applied to existing methods for CD/CF so that their impact on both problems can be jointly analyzed and visualized.
Specifically, we adapt \emph{Rehearsal}~\cite{rebuffi2017icarl} as one of the most common methods in CL to the federated setting. Our experiments show that the "Generalization Bump" that we found in the conventional Federated Continual setup still persists.

Obtaining these results even with such a simple method shows that the new unified contextualization towards both problems allows a holistic, joint analysis of methods for privacy-preserving improvements of robustness against spatio-temporal shifts.

\section{Related Work}\label{relatedwork}
In the following section, we present state-of-the-art algorithms for FL and give an overview of existing literature covering the analysis of CD. Following that, we investigate the current state of research on CL and discuss potential problems when training models sequentially, including the effect of CF. \emph{ Table 1} in the section \emph{Related Work} of the supplementary material gives an overview and categorization of methods for FL and CL. As the focus of this work is the joint analysis of CD and CF, we show the existing research on the analysis of these problems before we shed light on the limitations of current analysis that motivate this work.

\subsection{Federated Learning and Client Drift}
\emph{Federated Averaging (FedAvg)}~\cite{10} is the most commonly used FL algorithm and provides a basis for a wide range of other developments. There, the clients train locally and share their weights with the server, which averages them and pushes the resulting weight updates back to the clients.
~\citet{8} highlights the problem of CD by analyzing simple federated Stochastic Gradient Descent~\cite{SGD} on scenarios of population shift between federated clients on the datasets MNIST, CIFAR-10 and the KWS.
\textbf{Analysis of CD is mostly limited to the evaluation section of literature that introduces methods against it}.
Consequently, in the following, we give an overview of such contributions.
\emph{FedAvgM}~\cite{9}, adds \emph{momentum} to the updates of the server to reduce the effect of shifting clients. It is evaluated together with FedAvg on population shifts in CIFAR-10.
\emph{Scaffold}~\cite{11} follows a stronger approach by actively compensating CD, reducing the variance by control variates. It provides an evaluation against SGD and FedAvg on population-shifted MNIST. Another well-known approach is \emph{FedProx}~\cite{12}, which extends FedAvg by penalizing too large deviations of local weights from the global weights. It is compared to FedAvg on CD-population shifts to MNIST, FEMNIST, Shakespeare and Sent140.
\citet{li66supplementary} also only focuses on CD from population shifts on CIFAR-10 and ImageNet when comparing their contrastive learning-based approach to FedAvg, FedProx and SCAFFOLD.

\subsection{Continual Learning and Catastrophic Forgetting}
\emph{Continual Learning}~\cite{20,21} is a paradigm for learning on a stream of data instead of a closed dataset. A unified definition of when data becomes available or inaccessible varies widely between papers and authors. \emph{Catastrophic Forgetting}~\cite{19} occurs when the data distribution changes and the model adapts too strongly to the characteristics seen last, causing the performance of the model on previously trained tasks to decrease.

Similar to FL, the \textbf{most experiments on CF are provided in the context of evaluating original methods}. Therefore, the following provides an overview of such method papers that also perform experiments on CF:
~\citet{25} and ~\citet{26} mitigate CF by applying feature extracting and evaluate their methods on CF on various datasets. 
Similarly, besides evaluation against CF, ~\citet{27} applies fine-tuning of the parameters to make the model as discriminative as possible for the new task, while keeping the learning rate low to preserve the already learned structure.
The approaches from ~\citet{29} and ~\citet{30} decrease CF by adding additional nodes to the network when learning on new data.
The authors of \emph{Multitask learning}~\cite{28} keep a small subset of old train data to simultaneously train on them.
\subsubsection{Client Drift and Catastrophic Forgetting}
Most previous research is focused on either CD or CF, but there is also limited research that observed both problems:
Taking a look at the analysis of Continual Federated Learning, we find ~\citet{32} to first analyze CF in a non-federated setting, and then extend it to CF in FL. However, this analysis is limited to showing that CF also occurs in FL, and the work only covers the domain of Human Activity Recognition.
\citet{casado2022concept} parallelize CL and compare their approach to the conventional FL approaches FedProx and FedAvg in settings of CF on MNIST. However, this work does not cover CD.
\citet{yao2020continual} applies CL methods on the clients of FL to improve the initialization and the experiments on MNIST and CIFAR only focus on CF.
~\citet{33} propose a framework for Continual Federated Learning and analyze the performance decrease from CF on FedAvg, FedProx and their own proposed algorithm. But, similarly to other work, it does not analyze how the problems of CF and CD relate or interact. \textbf{This lack of existing research analyzing both problems jointly is a missed opportunity and the motivation for our work.}

\section{Method}
\begin{figure}[]
\centering
      \includegraphics[width=0.9\linewidth]{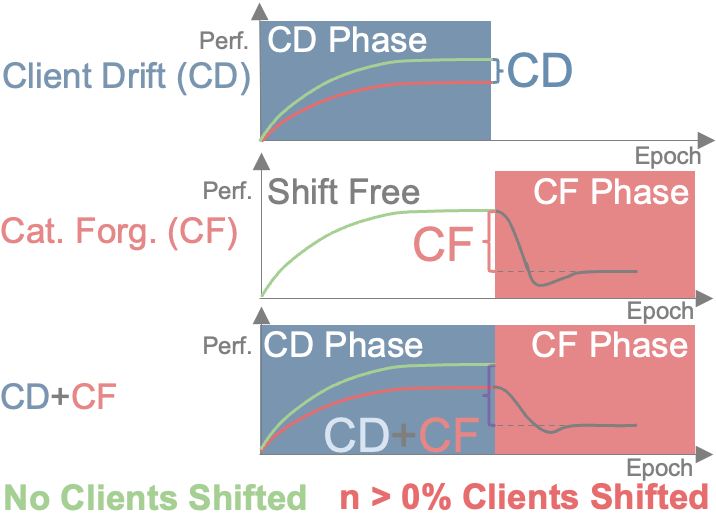}
      \caption{\textbf{CD} is measured by the perf. difference between shift free and a scenario with n perc. shifted clients. For \textbf{CF}, the clients are first trained without any shift, before switching to a given shift strength. CF results from the perf. drop since switching to shifted data. \textbf{CD + CF} results from first training \emph{n} shifted clients before shifting to a certain shift strength. CD + CF is the perf. difference to convergence in the unshifted scenario.}
      \label{fig:method}
\end{figure}
The goal of our framework is to simulate CD and CF to perform a correlation analysis between both problems and to investigate their interaction. In the following, we first describe our hypothesis on the correlation before explaining the actual experimental setup, which consists of the transformations to cause the shifts, followed by the setup for CD, CF and the combined CD/CF experiment.

\subsection{Problem Statement and Hypothesis for CD/CF correlation}
Our central thesis is that both CD and CF can be measured from a performance drop $\Delta p$.
${\Delta p}_{CD}$ from CD is caused when a certain ratio $\frac{s_{\hat{\alpha}}}{c}$ of clients with a client shift strength $\hat{\alpha}$ is shifted, where $s$ is the number of shifted clients and $c$, the total number of clients. Similarly, ${\Delta p}_{CF}$ from CF is caused by the data being perturbed through a certain shift strength $\alpha$.
The goal of this work is to investigate the relation between ${\Delta p}_{CD} = f\left(\frac{s_{\hat{\alpha}}}{c}\right)$ and ${\Delta p}_{CF} = g(\alpha)$.

We start by exploring the following hypothesis:
\begin{hyp}[H\ref{hyp:first}] \label{hyp:first}
The performance drop through Client Drift ${\Delta p}_{CD}$, caused by a certain share of shifted clients 
$\frac{s_{\hat{\alpha}}}{c}$ with a shift strength $\hat{\alpha}$, is correlated to the drop from Catastrophic Forgetting $\Delta p_{CF}$ that is caused by a certain shift strength $\alpha$.

\end{hyp}

Consequently, this means that in order to fulfill this hypothesis, we want to show correlation ($\sim$) between $\Delta p_{CD}$ and $\Delta p_{CF}$: $\exists m:{\Delta p}_{CD} \:{=} \:f( \frac{s_{\hat{\alpha}}}{c}){\sim}\; m\cdot g(\alpha) \:{=}\: m\cdot{\Delta p}_{CF}$.
This correlation is assumed to be piece-wise linear or possibly even fully linear.
We evaluate this on a dataset from natural images (CelebA) and a dataset from the medical domain (PESO).

\subsection{Simulation Framework}
We start by explaining how the framework causes shifts on both datasets in a controlled fashion by transforming the input data. We then outline our process for simulating CD and CF. Finally, we introduce our method for joint CD/CF analysis. Figure~\ref{fig:method} gives an overview of how all these types of shifts are measured.

\subsubsection{Transformations:} Both scenarios require emulating a shift of the training data with a defined strength. For each dataset, we use popular transformation methods for creating realistic shifts depending on the domain of the dataset.\\
We transform \emph{CelebA}, which consist of natural facial images, with the shifts proposed for \emph{ImageNet} by-C~\citet{hendrycks2019benchmarking}. These transformations provide the implementation of 15 types of image corruptions for natural photographs, which can be configured to 5 shift strength levels. The corruptions include various types of noise, blur, deformations and brightness- and contrast variations.\\
In the case of PESO from histopathology, which contains whole-slide images of prostate cancer, we use the transformations that are implemented in the digital pathology out-of-distribution detection framework \emph{FrOoDo}~\cite{stieber2022froodo}.
These \emph{transformations} by~\cite{schomig2021quality} are designed for histopathology slices and include brightness increase and adding overlays of thread, dark dirty spots, squamous epithelium or blood cells to the image.
Unlike the transformations on CelebA, which are applied at 5 different shift strengths, the transformations on the segmentation images on PESO can be freely configured by increasing the percentage of the impacted image.
Our detailed calibration process is described in the section \emph{Transformations on PESO} in the supplementary material.
\subsubsection{Client Drift Setup:}
In the following section, we explain how the framework can simulate the performance impact from CD, given a certain ratio of shifted clients $\frac{s_{\hat{\alpha}}}{c}$ so that we can relate this to CF and analyze their correlation (\textbf{H}\ref{hyp:first}) and their interaction.

We provide a \textbf{scenario with a fixed number of total clients}, and we can apply a \textbf{shift with a fixed strength} to each of them.
We choose \emph{FedAvg}~\cite{fedavg} as the method for FL, as it is the most widely used algorithm in the field.
First, we decide on a fixed number of global epochs for the training process based on the convergence behavior on the specific dataset and model.
We then train a federated model until convergence and record the global test accuracy for classification tasks and the global \emph{dice score}~\cite{dice1945measures} for segmentation tasks.
\textbf{We vary the percentage of shifted clients from 0 to 100 percent} with a step size of 10 percent so that, in total, 11 scenarios are covered.
We compare the final global Test Accuracy/Dice Score for each federation with the baseline with none of the clients shifted.
We get the performance drop from CD ${\Delta p}_{CD}$ depending on the ratio of shifted clients, which is used for the following analysis.

\subsubsection{Catastrophic Forgetting setup}
The goal is to simulate the impact of the shift strength $\alpha$ on the performance drop from CF ${\Delta p}_{CF}$.
In this scenario, \textbf{all clients experience the same shift strength $\alpha$}.
All clients are trained on ID data for the same number of epochs as before to ensure convergence. \textbf{After that pretraining, the train data is shifted by adding transformations with a certain strength to it}.
Again, we record the final performance metric on clean data.
The training continues for a fixed number of global epochs, which ensures convergence after the data shift. This allows us to compare the transformed federation with the federation trained without any transformations, giving us a quantification of the CF effect.
This experiment is repeated for all data shift strengths, allowing us to analyze the impact of the shift strength on the performance drop in an in-depth manner.

\subsubsection{Joint CD/CF analysis}
The framework offers the option to measure the joint performance impact of a certain level of CD and CF, which we use to investigate their interaction. Figure~\ref{fig:method} shows how we combine and measure CD and CF: First, we train the network in the previously introduced federated setup until convergence with a given ratio of shifted clients. We then plot the resulting performance difference between the convergence performance without any shifts and the convergence performance after running the CD and the CF experiment in a 3D plot.
The axes of this plot are: The ratio of shifted clients (CD), the shift strength (CF) and the performance change.
This allows us to visualize how a combination of CD and CF impacts the performance of the central model.

In this section, we first describe our experimental setup and implementation details in a comprehensive manner. 
First, CD and CF are analyzed in an isolated environment. From these, we conduct our shift relation analysis separately for each dataset. We then leverage the joint analysis capability of our framework to analyze their interplay by evaluating the performance for combinations of mixed CD and CF. Finally, we showcase the ability to analyze existing methods regarding both shifts with the example of the CL method rehearsal applied in a federated continual setting.
For better reproducibility, the section \emph{Experiment Details} in the supplementary material provides pseudocode of the experiments.

\subsection{Setup and Parameters}

\subsubsection{Datasets}

We use two datasets from different domains:

\emph{CelebA}~\cite{liu2015faceattributes} consists of \textbf{202,599 images} of resolution \textbf{178x218}, showing the faces of 10,177 celebrities.
Besides 5 landmark locations, each image is annotated with 40 binary attributes.
We focus on the binary attribute of \emph{smiling} for a consistent impact from shifts, which is not class-dependent.

In the medical domain, we use the \emph{PESO} dataset for Prostate Epithelium Segmentation. The data consists of \textbf{102 whole slide images} (WSIs) that show tissue and were originally scaled at 0.48 \textmu m/pixel. In order to decrease the computational effort, we downsample them by factor 4, resulting in a scale of 1.92 mu/pixel and a resolution of \textbf{11,871x26,323} pixels.
We assign each client a set of individual WSIs for training, which sample patches of \textbf{114x114} pixels from the annotated regions of interest.

\subsubsection{Implementations and Parameters}
On \textbf{CelebA}, we train a \textbf{Vision Transformer} for the classification problem, as introduced by ~\citet{vit}.
For the Vision Transformer, we use Stochastic Gradient Descent with a batch size of 32, a learning rate of 0.02, and Cross Entropy loss.
As a FL algorithm, we deploy the \emph{FedAvg} implementation of ~\citet{vitfl}, with the number of \textbf{local epochs set to 1} and \textbf{227 clients}.
We train for \textbf{200 global communication rounds} to achieve convergence, and in the case of the CF experiment, we train for \textbf{100 additional global communication rounds} after shifting the train data.
For all experiments on CelebA, we run \textbf{10 seeds}.

For the evaluation on \textbf{PESO}, we use the implementation of a \textbf{U-Net}~\cite{ronneberger2015u} in \emph{BottleGAN}~\cite{bottlegan}.
BottleGAN is a framework that provides implementations of federated algorithms such as \emph{FedAvg}~\cite{fedavg} on a U-Net for WSI segmentation tasks.
The implementation utilizes the Adam optimizer with $\beta_1=0.9$ and $\beta_2=0.999$ and a learning rate of 0.001. We apply Cross Entropy loss and a batch size of 48.
As the algorithm, we implement \emph{FedAvg} for FL in BottleGAN with \textbf{local epochs set to 1} and \textbf{10 clients}.
We train for \textbf{500 global communication rounds} to achieve convergence. In the CF setting, we train for \textbf{40 additional epochs} after shifting the train data.
Due to the higher computational effort of training BottleGAN on PESO, we repeat each experiment for \textbf{3 seeds}.
More details for reproducing the setup for both datasets are attached in the supplementary material in \emph{Experiment Details}.
\section{Experiments}
\begin{figure*}[ht] 
  \begin{minipage}[b]{0.5\linewidth}
    \centering
    \includegraphics[width=0.7\linewidth]{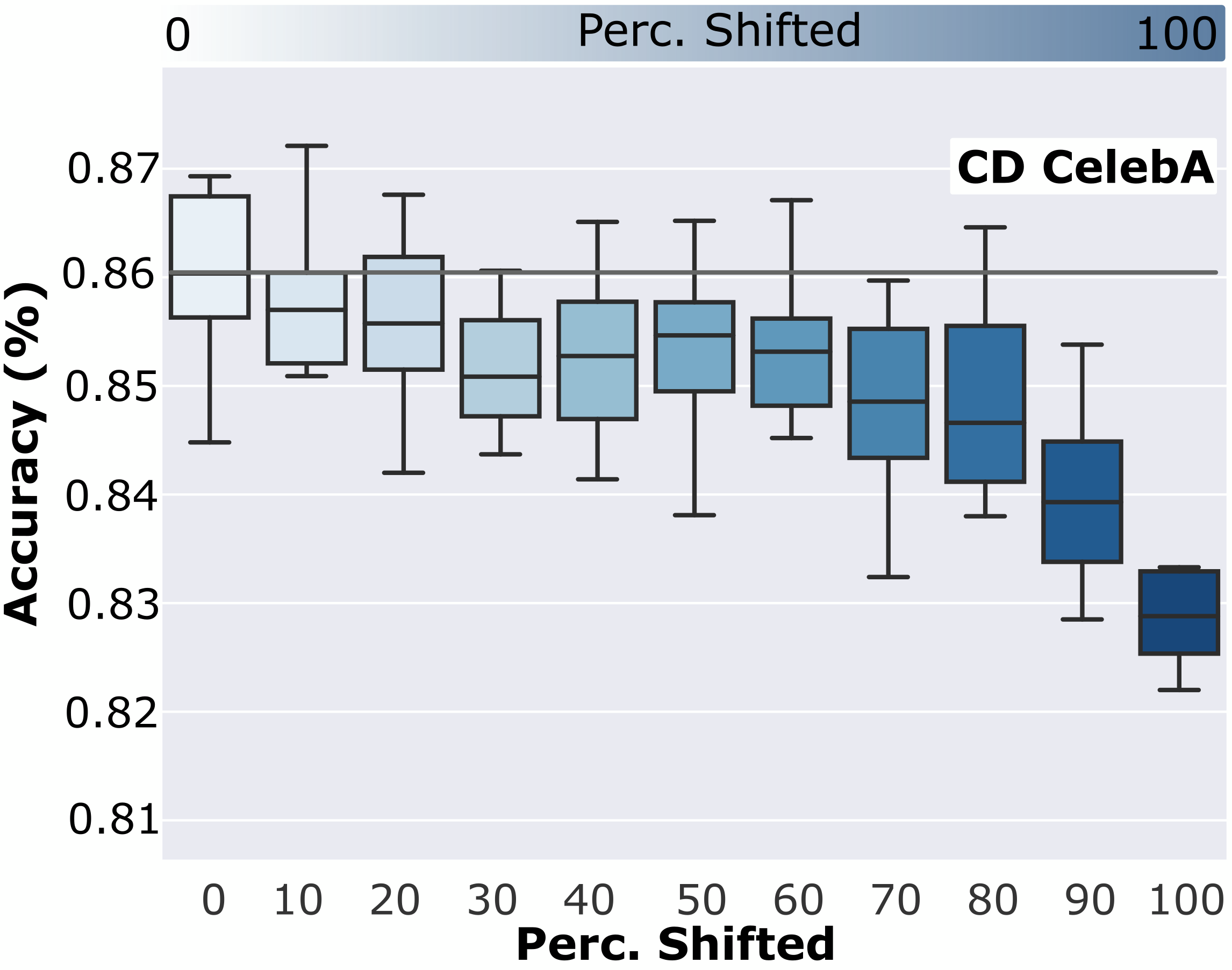} 
    \caption{Test Acc. on \textbf{CelebA} for shifted clients \textbf{(CD)}} 
    \label{fig:cdceleba}
  \end{minipage}
  \begin{minipage}[b]{0.5\linewidth}
    \centering
    \includegraphics[width=0.7\linewidth]{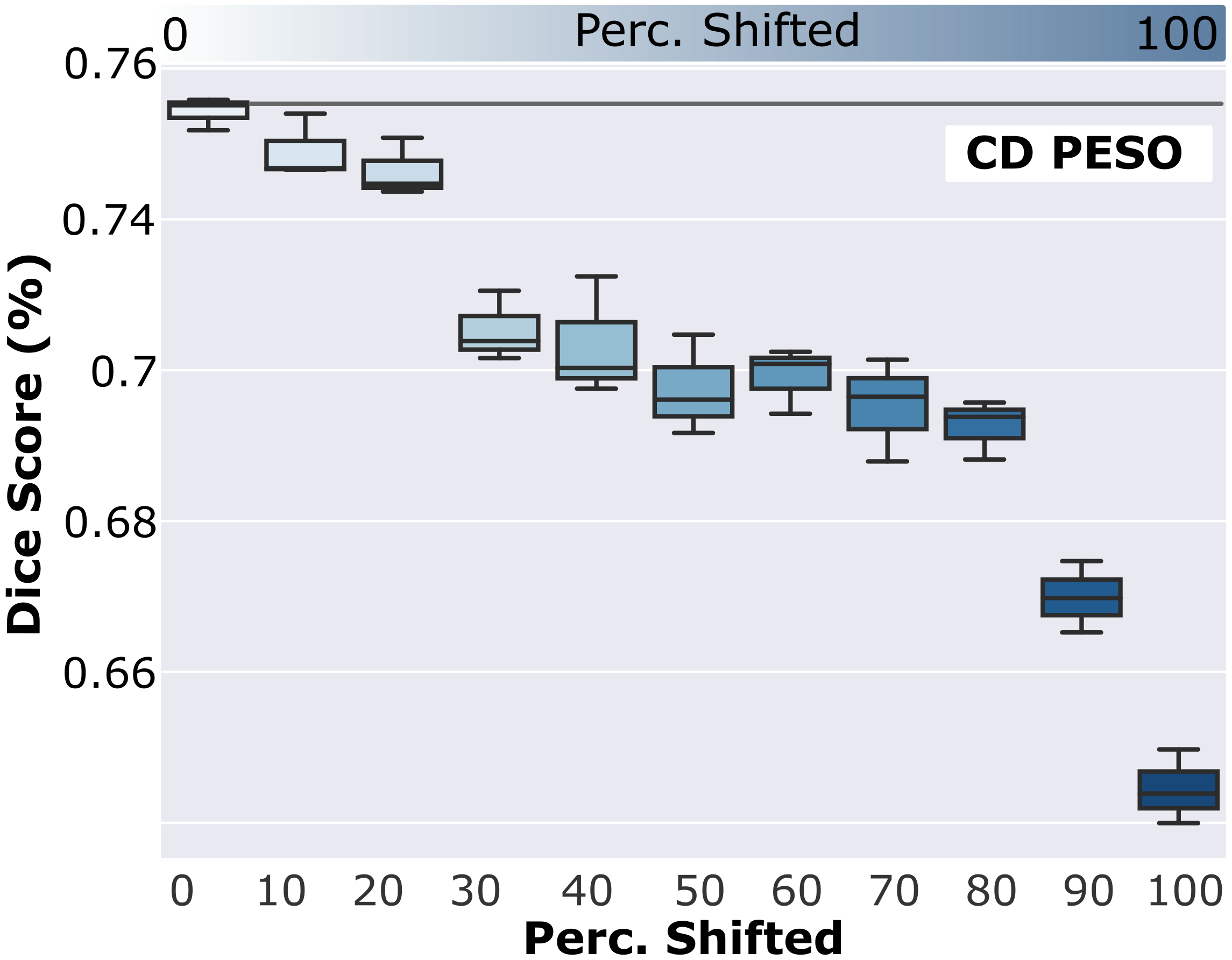} 
    \caption{Dice Score. on \textbf{PESO} for shifted clients \textbf{(CD)}}
    \label{fig:cdpeso}
  \end{minipage} 
  \begin{minipage}[b]{0.5\linewidth}
    \centering
    \includegraphics[width=0.7\linewidth]{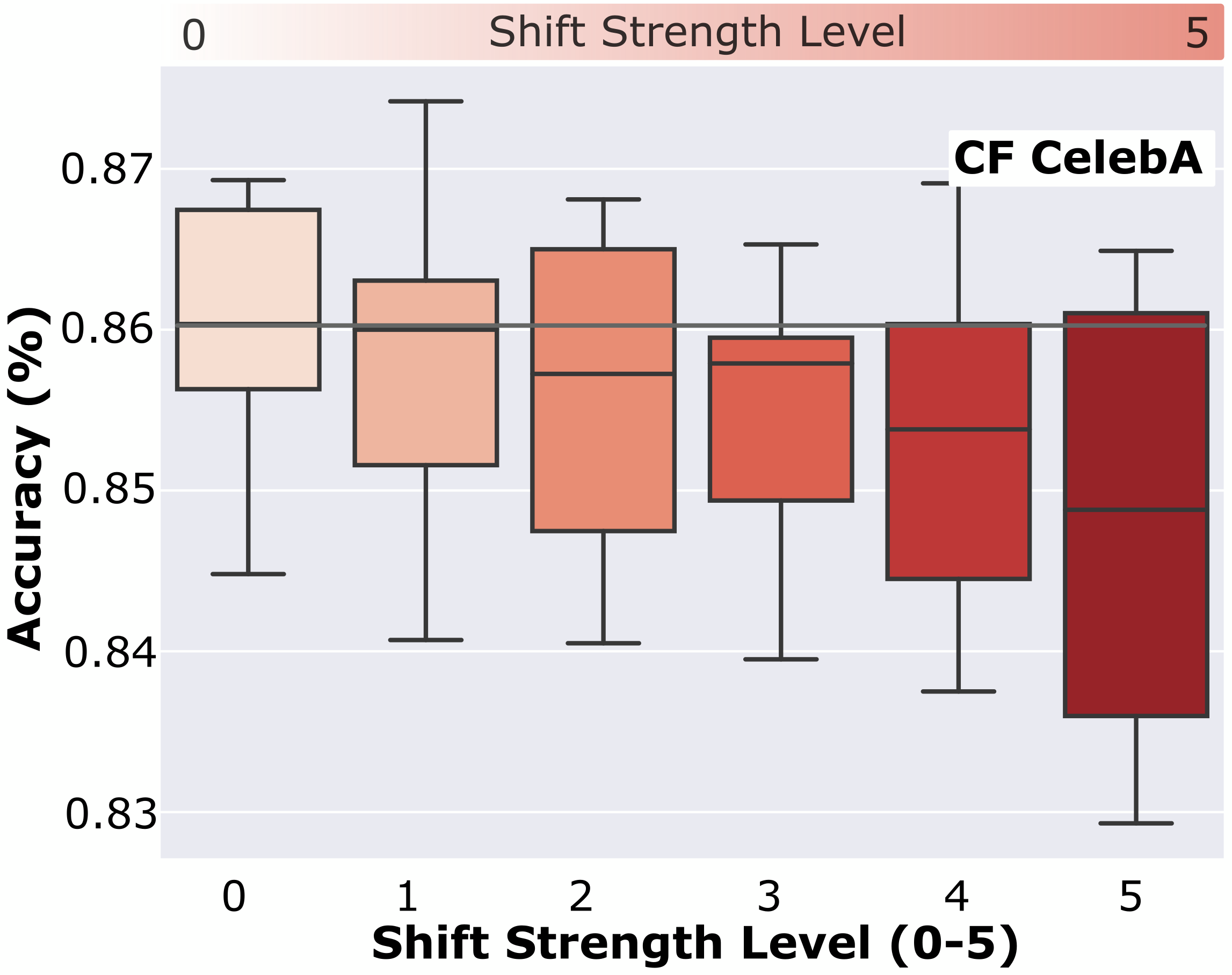}
    \caption{Test Acc. on \textbf{CelebA} for \textbf{CF} from shift strengths}
    \label{fig:cfceleba}
  \end{minipage}
  \begin{minipage}[b]{0.5\linewidth}
    \centering
    \includegraphics[width=0.7\linewidth]{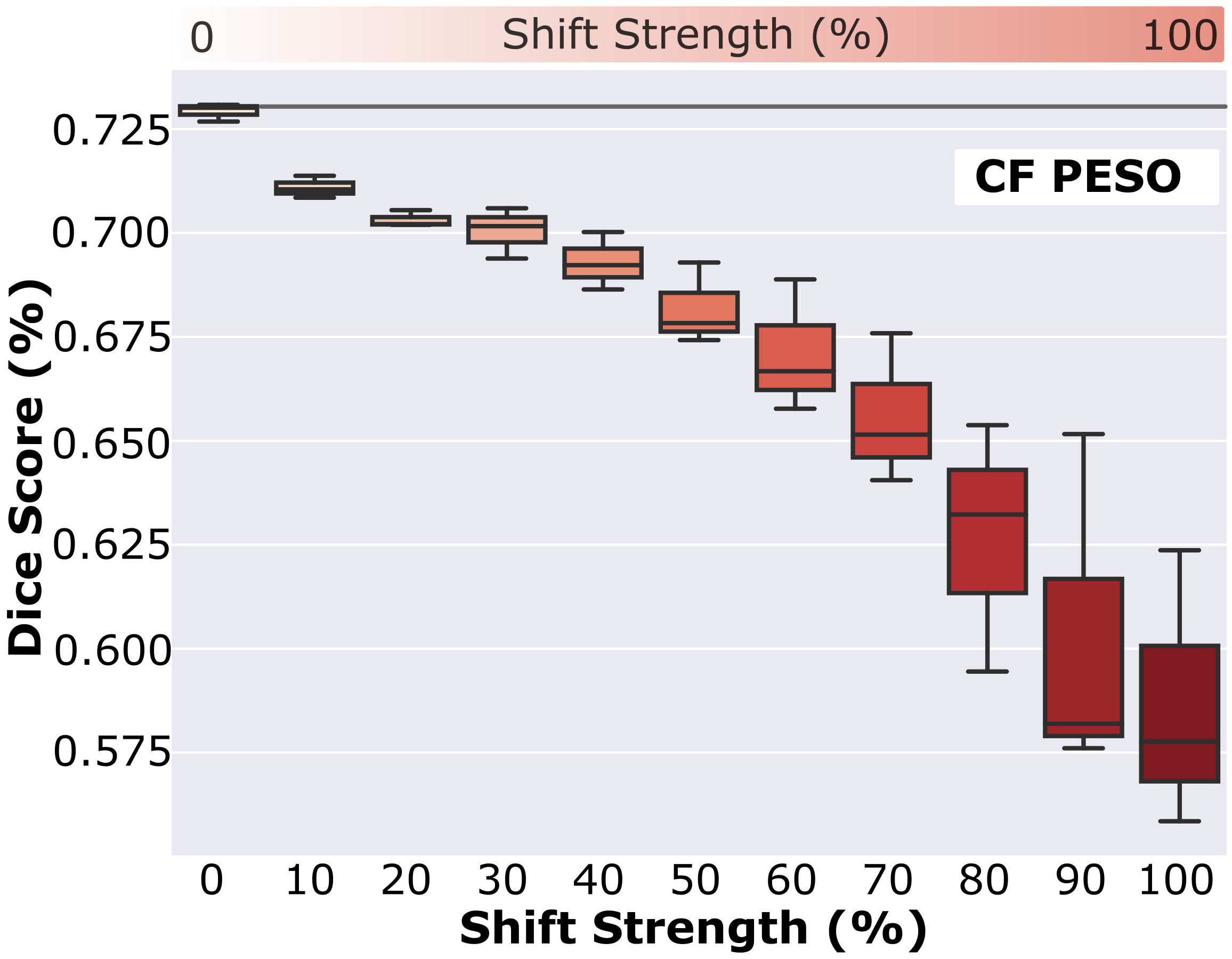}
    \caption{Dice Score on \textbf{PESO} for \textbf{CF} from shift strengths}
    \label{fig:cfpeso}
  \end{minipage} 
\end{figure*}
\subsection{Client Drift}\label{sec:cd}
In this experiment, we analyze how the ratio of shifted clients affects the performance drop through CD. Later we combine these with the results from CF for the analysis of their correlation.

Applying the aforementioned experiment for CD on CelebA leads to noticeable performance degradation during the whole training process. Figure~\ref{fig:cdceleba} shows the degraded final test accuracy on CelebA with different numbers of shifted clients contributing to the central model.
With each step increasing the ratio of shifted clients, the absolute accuracy on ID test data decreases.
We plot the relative accuracy drop compared to the shift-free \textbf{(ID)} scenario as a line. This line indicates similar results, as it increases with a larger ratio of shifted clients. 

In Figure~\ref{fig:cdpeso}, we see that the results of this CD setting on PESO are similar regarding the relationship between the ratio of shifted clients and the impact on the ID data test accuracy. Both datasets implicate the same tendency. We also observe that, for both datasets, the performance drops more severely when at least $\frac{2}{3}$ of the clients are shifted. However, the strength of the performance drop is different. In the scenario with all clients shifted, the relative performance drop in ID data test accuracy on CelebA is only 3.44\%, while the Dice Score on PESO drops by 11.89\%.
The difference can be explained by robustness differences of the underlying domain-specific models, as shown by~\citet{qu2022rethinking}.
Furthermore, in a classification task like CelebA, other less affected image parts can compensate for a correct prediction. In contrast, in the segmentation setting, lost information on images tends to degrade the segmentation performance in the corrupted image regions severely.
Nonetheless, the performance trend is similar in both datasets, which shows that the CD effect can be observed in both settings.

\subsection{Catastrophic Forgetting}\label{cfresults}
After analyzing CD, we analyze how the shift strength in a CL scenario affects the performance drop through CF. We seek these results to combine them with the results from CD and show their correlation.
When switching to the shifted data during the second part of CL on CelebA, the ID data test accuracy on the initially trained uncorrupted data drops in the first epochs. The performance remains decreased for the whole remaining 100 training epochs. The final ID data test accuracy on CelebA is shown in Figure~\ref{fig:cfceleba}.
An increased shift strength decreases the accuracy on the ID test data. The gradient of the relative performance drop is the highest when the shift strength is at over $\frac{2}{3}$. This observation is similar to the CD scenario, as the shift strength relates to the ratio of shifted clients, as in subsection~\emph{Client Drift}. We discuss these similarities in more detail in the section~\emph{Relationship Analysis}.

Applying our scenario of CF shows a similar behavior on PESO in Figure~\ref{fig:cfpeso} as the performance drop from an increasing shift strength level is also present here. Moreover, the federated systems show the same behavior of faster performance degradation in the last third of its shift strength.
As in the CD setting, the performance drop on PESO is significantly higher, with up to 18.89\%, compared to CelebA, with up to 1.24\%. The possible reasons for this are similar to the ones explained in section~\emph{Client Drift} and relate to domain-specific behavior as well as model architecture and the classification vs. segmentation task on hand.
An additional explanation, in the case of CF, is that some of the shifts utilized for CelebA, are implicitly covered by the initial training distribution. Images with natural variations from the acquisition process, e.g., varying flashlights and devices as in CelebA are different to the WSIs from PESO that are acquired in a more controlled way by a scanner.

\subsection{Relationship Analysis}\label{Relationship Analysis}
\begin{figure}[ht] 
  \label{ fig7} 
  \begin{minipage}[b]{0.5\linewidth}
    \centering
    \includegraphics[width=0.9\linewidth]{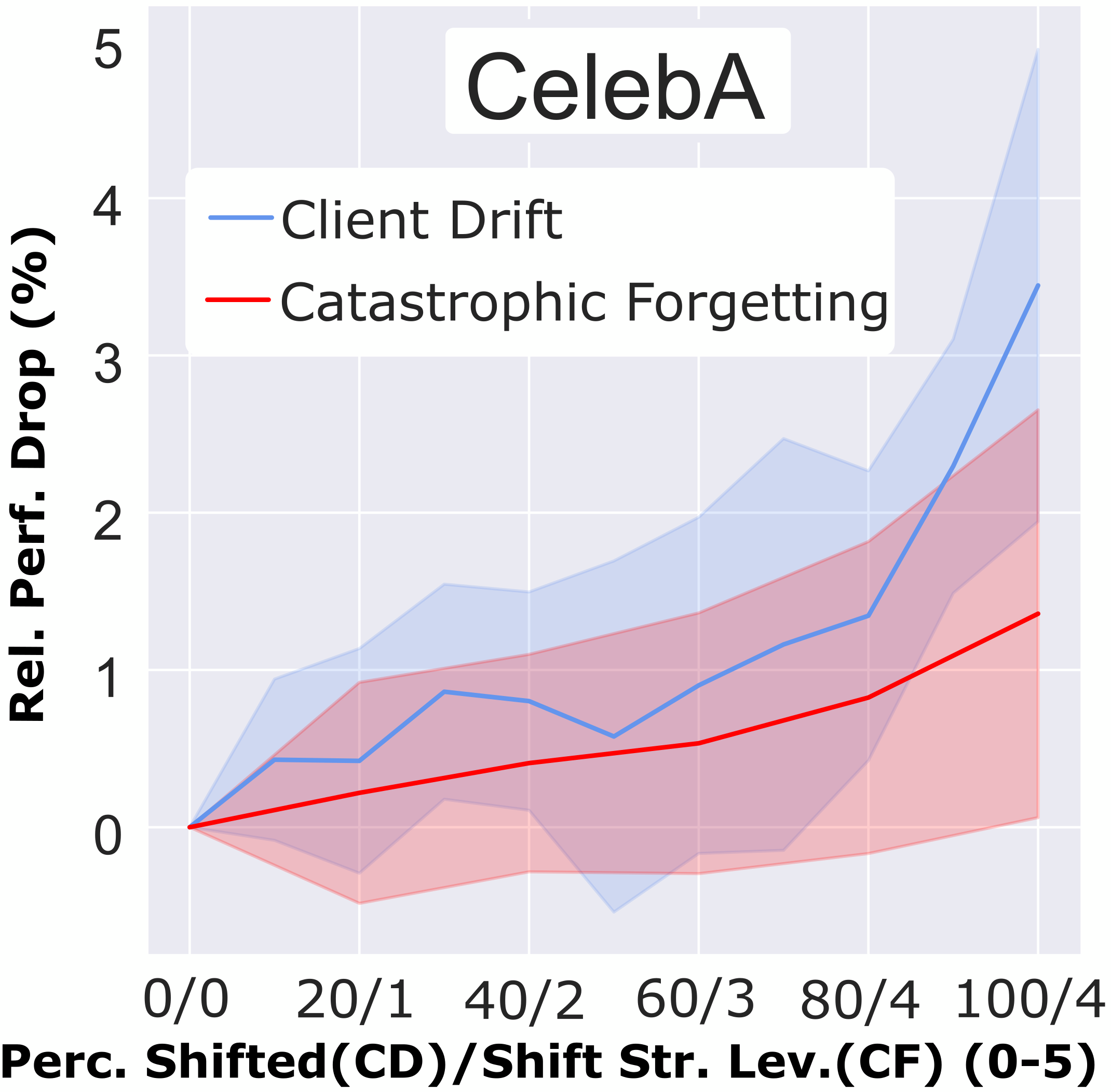}
  \end{minipage}
  \begin{minipage}[b]{0.5\linewidth}
    \centering
    \includegraphics[width=0.9\linewidth]{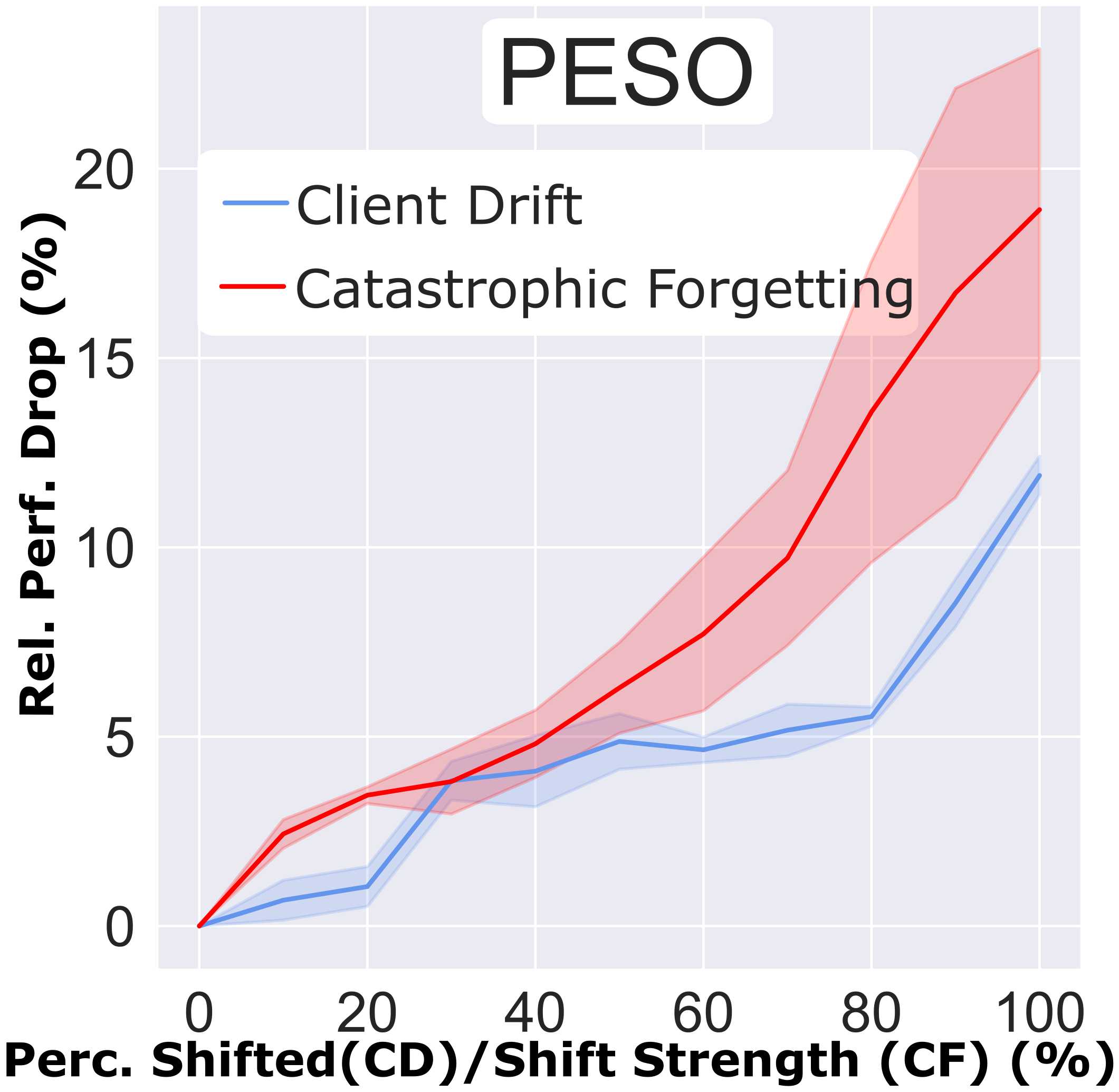}
  \end{minipage} 
\caption[width=0.5\linewidth]{Comparison of performance drop from CD vs. CF}
\label{fig:correlation}
\end{figure}
For investigating \textbf{H}\ref{hyp:first}, we visualize the relation between $\frac{s_{\hat{\alpha}}}{c}$ and $\Delta p_{CD}$ compared to the relation between $\alpha$ and $\Delta p_{CF}$.

For this, we plot the ratio of shifted clients in the CD scenario and the relative shift strength of the CF scenario on the x-axis. The y-axis indicates the relative performance drop from both problems individually.
We visualize this in Figure~\ref{fig:correlation} as two similar plots due to the individual nature of both datasets.
To increase the comparability of all results, we linearly interpolate the five shift strength levels on CelebA.
Both resulting plots for PESO and CelebA clearly show a similar trend that the ratio of shifted clients increases the CD, and the shift strength continuously increases the CF. \textbf{In most parts, both plots for CD and CF are highly correlated}, which supports \textbf{H}\ref{hyp:first}.
The \textbf{Spearman's correlation of 1.0 on CelebA and 0.9909 on PESO} strongly supports the assumption of a piece-wise linear correlation, while the \textbf{Pearson correlation of 0.9712 on CelebA and 0.9385 on PESO} indicates even an almost fully linear correlation. This strong correlation clearly supports \textbf{H}~\ref{hyp:first} and motivates to see CD and CF as a related problem.

\subsection{Spatio-temporal Generalization Analysis}\label{stanalysis}
\begin{figure}[ht]
\centering
      \includegraphics[width=0.9\linewidth]{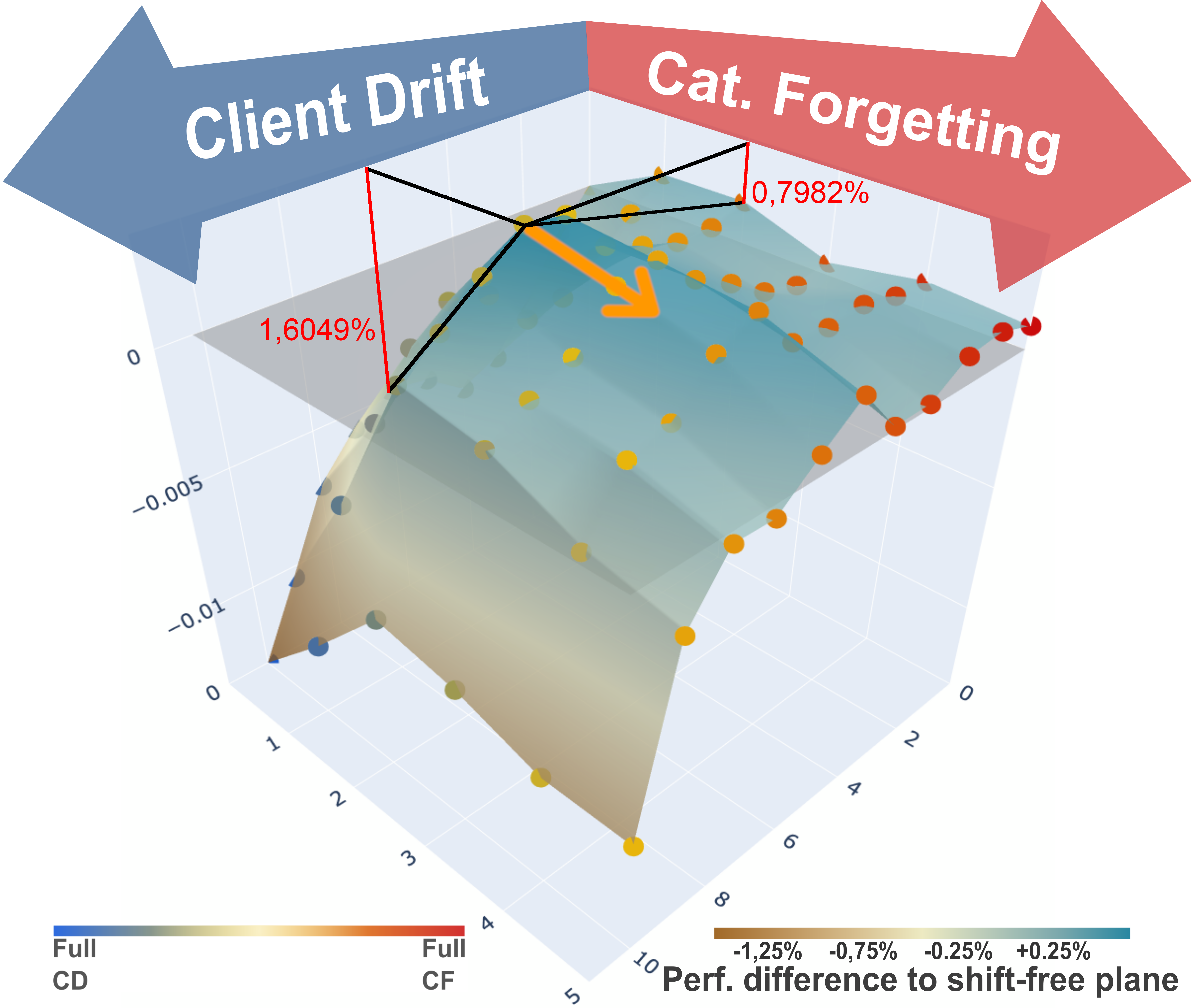}
      \caption{CD + CF visualization with \textbf{rehearsal} on CelebA. The orange arrow shows how the generalization bump is shifted towards stronger CF when combined with CD. The red lines and percent numbers indicate, for the peak, how the performance in combined CD+CF improves compared to only applying the respective shift individually (CD/CF).}
      \label{fig:landscape_reh}
\end{figure}
To obtain a better understanding of the relationship between CD and CF and how they influence each other, we also visualize mixed scenarios with concurrent CD and CF at various levels on CelebA in Figure~\ref{fig:mesh1}.
This shows the same correlating behavior as discussed before. Besides that, the combination of moderate CD and moderate CF can even slightly increase the performance. 
In the center of Figure~\ref{fig:mesh1}, we see a phenomenon we interpret as \textbf{"generalization bump"}. The spatio-temporal variations for the federation force the model to a better generalization when moderate CD and CF are combined. We exemplarily measure this for the peak of this generalization bump by comparing how the performance is improving compared to having only one shift (CD or CF) as indicated by the red lines. The red numbers represent the relative performance improvement compared to only applying the respective shift (CD/CF) individually. Specifically, the \textbf{performance improves by 0.6407\% when adding CD to the corresponding CF scenario. Adding CF to CD has a larger impact on the generalization, as it improves performance by 1.4012\%.} However, when at least one factor (CD or CF) increases beyond this peak, the performance decreases again, and the federation can not compensate with a better generalization.
The observation that increased CD, as well as CF, can both lead to this result further supports \textbf{H}\ref{hyp:first}.

To amplify the ability of our joint analysis approach, we apply it to rehearsal, one of the most common methods for CL. We adapt this method to work in the federated setting. Details of these adaptations as well as a performance evaluation after using our framework can be seen in the supplementary material in the section \emph{Rehearsal}.
After simulating the same combination of CD/CF on rehearsal, we plot it again in Figure~\ref{fig:landscape_reh}. It shows that even with rehearsal, the generalization bump is still present, which supports that this is a reoccurring phenomenon. As in the conventional scenario, adding CF to the CD scenario causes the largest performance improvement with 1.6049\% while adding CD to CF still improves it by 0.7982\%. Moreover, it shifts towards stronger levels of CF in combination with CD. This highlights the importance of a joint approach for CD and CL that is evaluated in a joint analysis to improve robustness for spatial- and temporal shifts concurrently.

\subsection{Ablation Study}
In order to gain a better understanding of the impact of the 15 specific transformations provided by \emph{Imagenet-C}~\cite{hendrycks2019benchmarking} on CD and CF, we evaluate them separately.
The results are attached in the supplementary material in the section \emph{Transformations on CelebA}.

\section{Conclusion and Future Work}
Federated and Continual Learning allow the training of Deep Learning models in dynamic real-world settings while protecting data privacy. They were designed to address, respectively, spatial and temporal restrictions on data availability. The main challenges to overcome in order to obtain robust performance are Client Drift and Catastrophic Forgetting.
Existing work that explores CD and CF offers separate analyses of only one problem, hindering the development of methods for a robust deployment under spatio-temporal shifts.
Motivated by this, \textbf{we introduce a framework for joint CD/CF analysis}.
To show the potential of a joint approach, we analyze the mechanics of CD and CF as the relation between the ratio of shifted clients for CD and the shift strength for CF. We show the correlation between both problems and support this novel view by showing that \textbf{the average Pearson rank correlation coefficient between CD and CF is over 0.9712 on \emph{CelebA} and 0.9385 on the histopathology dataset \emph{PESO}}.
To further leverage the joint analysis capabilities of our framework, we simulate mixed CD/CF and plot it for the analysis of their interaction, both with and without rehearsal, as an existing method against CF. This analysis shows an interesting, reoccurring phenomenon of a \textbf{performance improvement when moderate CD and CF are combined (generalization bump.)}

\bibliography{aaai24}

\newpage
\section{Supplementary Material}
\subsection{Transformations on CelebA}
\subsubsection{Ablation Study}
\begin{figure}[h]
    \centering
    \includegraphics[width=1.0\linewidth]{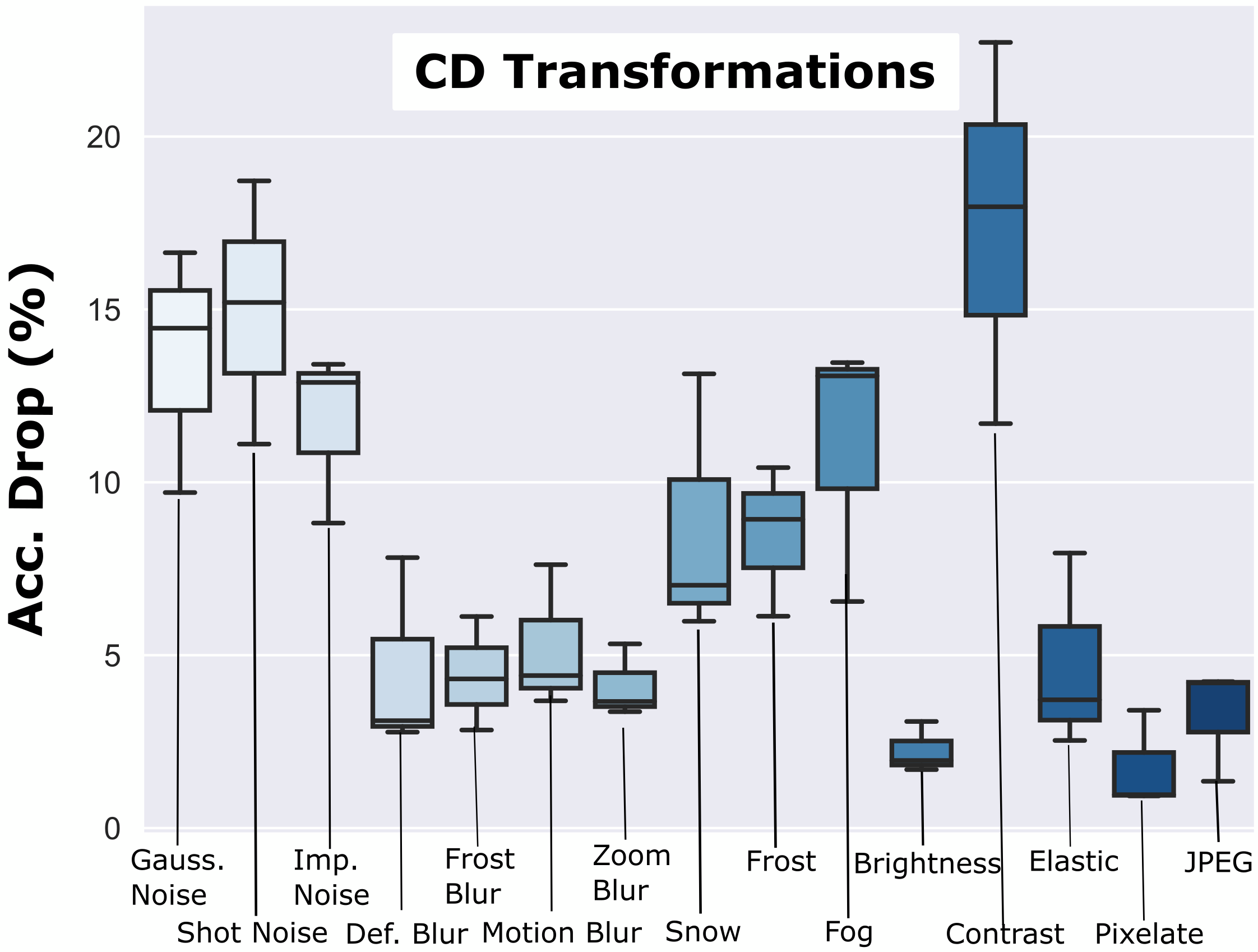}
    \caption{Relative performance drop from the specific transformations in the CD Experiment of CelebA}
    \label{fig:ablationaugs_cd}
\end{figure}

\begin{figure}[]
    \centering
    \includegraphics[width=1.0\linewidth]{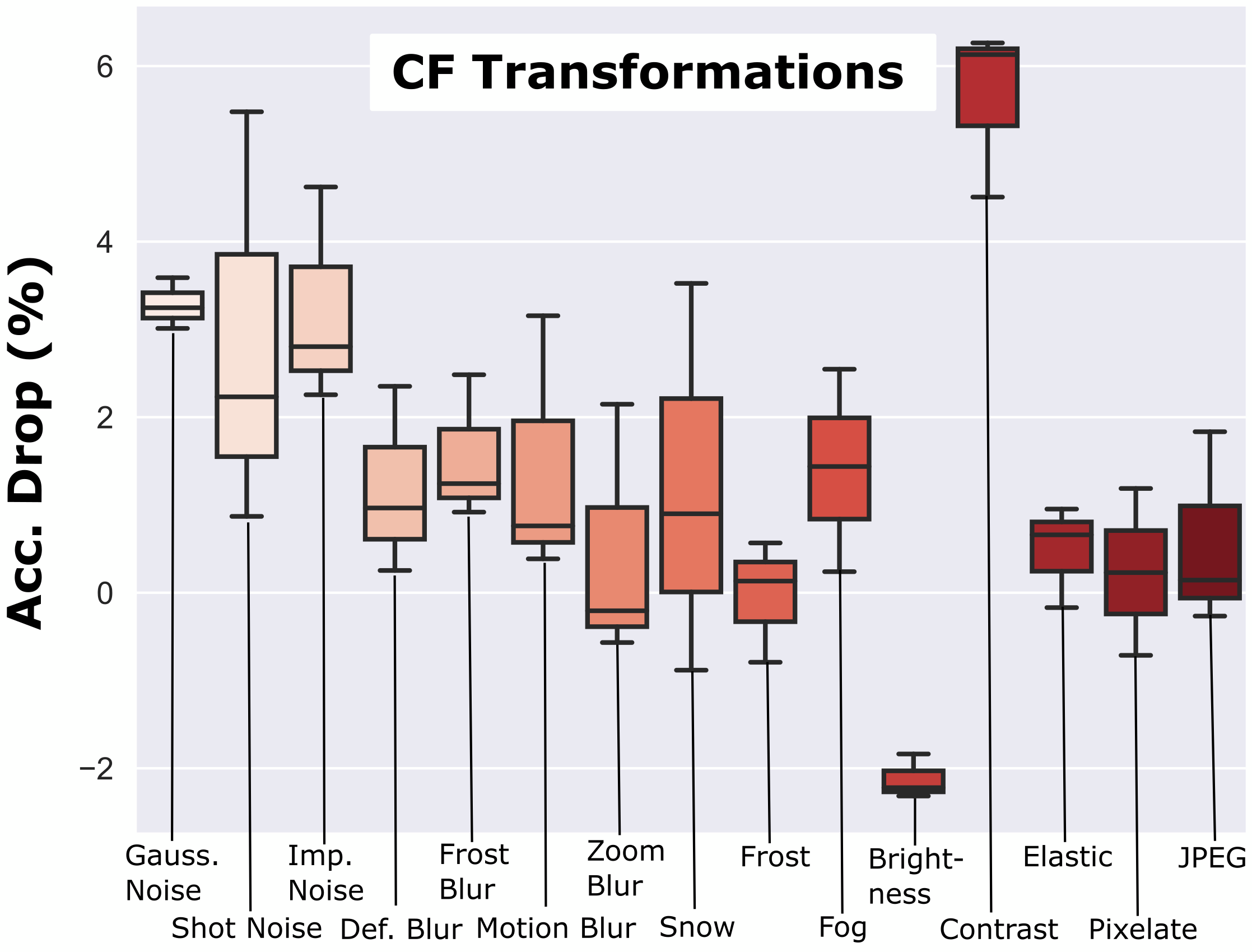}
    \caption{Relative performance drop from the specific transformations in the CF Experiment of CelebA}
    \label{fig:ablationaugs_cf}
\end{figure}

\begin{figure}[]
    \centering
    \includegraphics[width=0.4\textwidth]{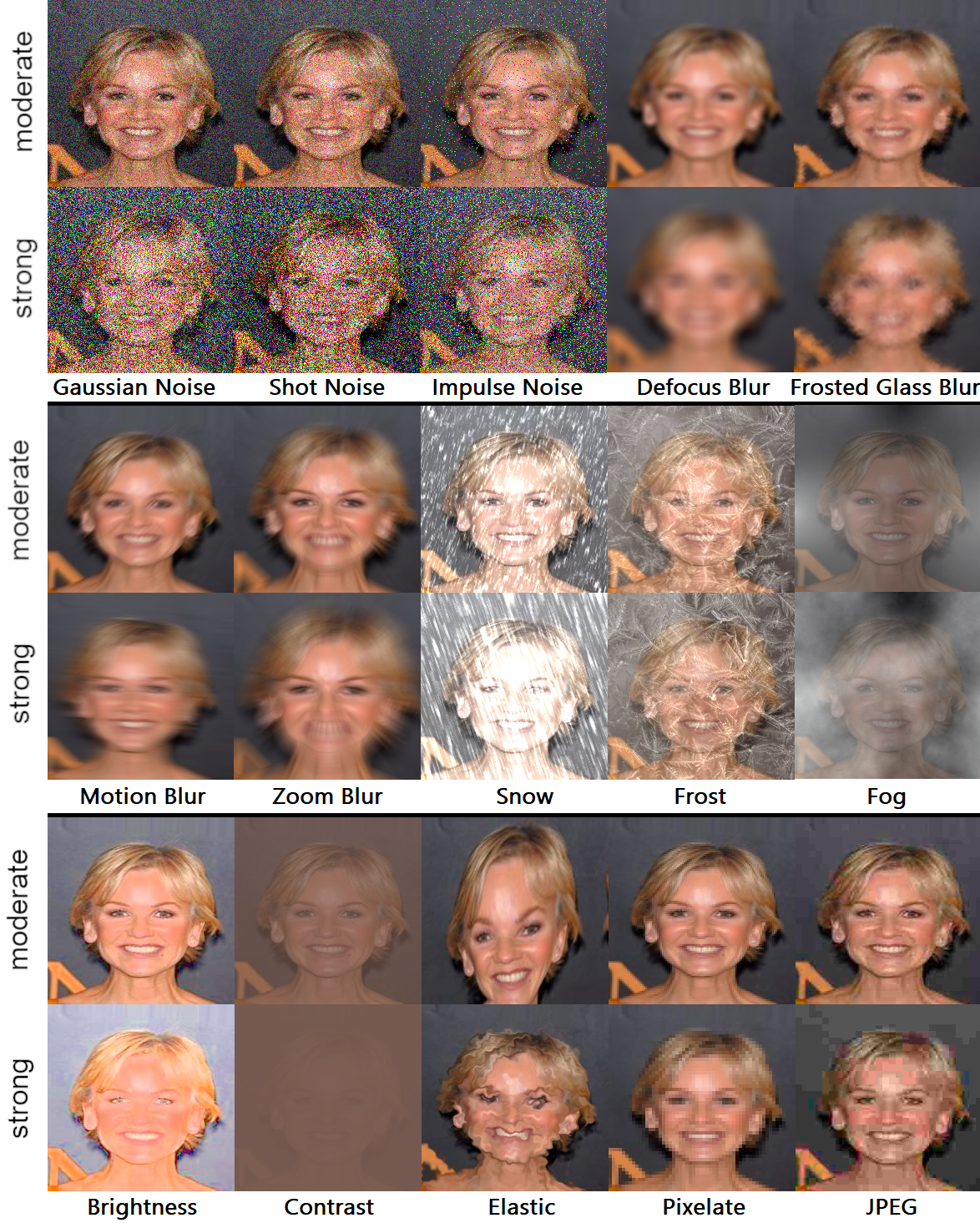}
    \caption{Sample of the transformations applied to the CelebA dataset in moderate and strong severity levels}
    \label{fig:CelebA_transforms}
\end{figure}

In the following, we investigate how the given transformation types for CelebA impact CD and CF.
We perform the usual experiments for each transformation type individually at full strength and in both scenarios but with fewer epochs due to the large computational effort.
We train for \textbf{100 global epochs} in the CD scenario and \textbf{50 additional epochs} in the CF scenario.
~Figure~\ref{fig:ablationaugs_cd} visualizes the resulting relative accuracy drop in CD and Figure~\ref{fig:ablationaugs_cf} in CF compared to the shift-free baseline.
It shows that the contrast transformation has the largest impact in both scenarios. As the goal is to classify if the person is smiling, a change in contrast, can reduce the visibility of the mouth in the face, which explains why this degrades the performance. Moreover, noise transformations have a large impact as they disturb the ability to detect the structure of the mouth.
Figure~\ref{fig:CelebA_transforms} shows each transformation type on a sample image from CelebA at moderate as well as strong severity levels.

\subsection{Transformations on PESO}
\subsubsection{Calibration}
To ensure that the transformations are usable for a consistent emulation of distribution shifts, we perform the following calibration process.

For the overlay-based transformations that add artifacts, such as the thread on top of the existing image, we set the transformation severity with the affected image area scaling to ensure that a certain share of the image is covered. Moreover, we configure the opacity to control what details remain visible under the overlay. In the case of the brightness transformation, we configure the severity by changing the brightness factor. 

\begin{figure}[H]
    \centering
    \includegraphics[width=0.45\textwidth]{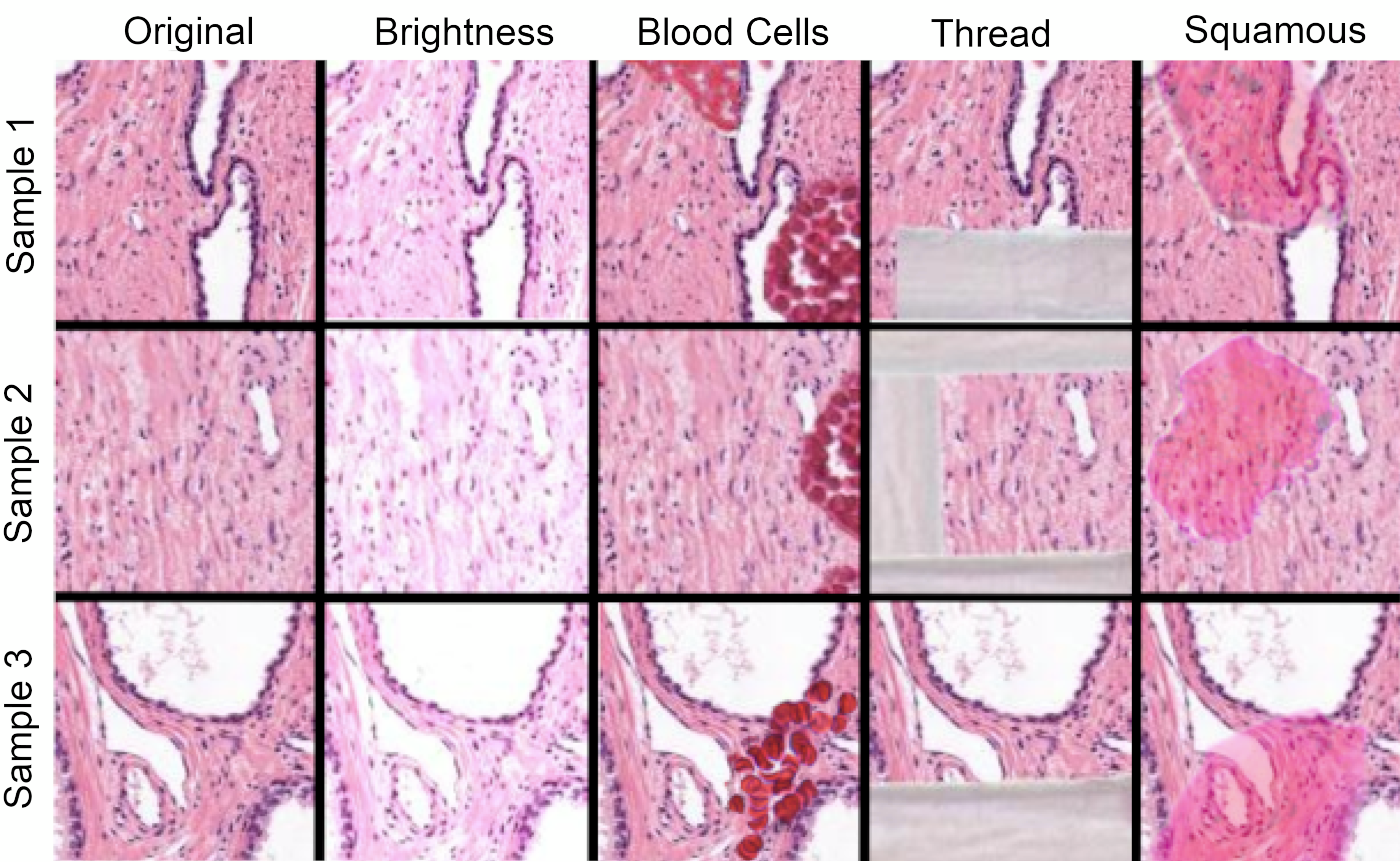}
    \caption{Samples of the transformations applied to the Whole Slide Images from the PESO dataset}
    \label{fig:PESO_transforms}
\end{figure}

In the first calibration step, we tune these parameters in such a way that the Dice Score on our pretrained standard federation drops by 20\% from $94\%$ on clean test data to $75\%$ with the respective transformation. After that, we perform two more tests to ensure that these transformations are feasible to emulate CF. We first apply the resulting transformation to train the same federation for the same 500 epochs required to achieve convergence. The final model is tested on clean data, and we confirmed that the Dice Score drops by at least $5\%$ from this distribution shift. In the last step, we test for the ability to emulate CF. Therefore, we test the clean pretrained model separately on clean and transformed data. At the beginning of the distribution shift, we expect a higher Dice Score on clean data than on transformed data. In the following, we expect a changing order. If this is the case, this indicates CF. We perform all the stated tests with the transformations used in this work to ensure their feasibility for consistent distribution shifts. Figure~\ref{fig:PESO_transforms} shows samples of the resulting transformations on PESO.

\subsection{Related Work}
\begin{table*}[]
\begin{center}
\begin{tabular}{|c|c|c|}
\hline
\textbf{Problem} & \textbf{Meth. Category} & \textbf{Method} \\ \hline
\multirow{9}{*}{} & Penalizing \multirow{2}{*}{} & \emph{FedAvgM}~\cite{9}  \\ \cline{3-3} 
                  & Outliers                     & \emph{FedProx}~\cite{12} \\ \cline{2-3} 
                  & Compensate                   & \emph{Scaffold}~\cite{11} \\ \cline{2-3} 
                  & \multirow{5}{*}{}            &  \citet{hinton2015distilling}\\ \cline{3-3} 
 Client Drift     &     Knowledge                &  \emph{FedDF}~\cite{14}\\ \cline{3-3} 
       (CD)       &     Distillation             &  \emph{FedMD}~\cite{15}\\ \cline{3-3} 
                  &         (KD)                 &  \emph{CRONUS}~\cite{16}\\ \cline{3-3} 
                  &                              &  \emph{FedAUX}~\cite{17}\\ \cline{2-3} 
                  & Data Sharing                 & \emph{FedMatch}\cite{jeong2021federated} \\ \hline
\multirow{8}{*}{} & \multirow{3}{*}{}            &  \emph{Donahue et al.}~\cite{25}\\ \cline{3-3} 
                  & Param. Tuning                &  \citet{26}\\ \cline{3-3} 
                  &                              &  \emph{Girshick et al.}~\cite{27}\\ \cline{2-3} 
 Catastrophic     & \multirow{2}{*}{}Network     &  ~\citet{29}\\ \cline{3-3} 
 Forgetting       &            Extension                  &  \citet{30}\\ \cline{2-3} 
      (CF)        & \multirow{3}{*}{} &  \emph{Multitask Learn.}~\citet{28}\\ \cline{3-3} 
                  &        Keep data/logits         &  \emph{Rehearsal}~\citet{rebuffi2017icarl}\\ \cline{3-3} 
                  &                               &  \cite{31}\\ \hline
\end{tabular}
\end{center}
\caption{Overview of the existing Federated and Continual Learning methods tackling, respectively, Client Drift and Catastrophic Forgetting.}
\label{tab:relatedwork}
\end{table*}
In Table \ref{tab:relatedwork} we give an overview of existing approaches that address CD/CF individually and categorize them.
We start with a short introduction of the main categories from the table for approaches against CD:
One direction is to either just \emph{penalize outliers}~\cite{12} while other methods go further by \emph{actively compensating CD}~\cite{11}.
Alternatively, sharing a global dataset enables a better generalization and makes the model less prone to CD~\cite{jeong2021federated}.
A very different approach is to apply \emph{Knowledge Distillation}~\cite{hinton2015distilling}:
First, the clients train their local model. After that, they teach the server by providing their output logits. Besides making the global model more flexible to different inputs, this intermediate representation step reduces CD.
Beyond these FL-related methods, there are separate approaches against CF in CL:
One direction is \emph{parameter tuning}~\cite{25} to balance learning new data distributions while preserving the existing knowledge.
Alternatively, some approaches \emph{extend the network}~\citet{30} to keep distribution-specific weights.
\emph{Keeping data/logits}~\citet{rebuffi2017icarl} is a widely used approach, which we also apply in our experiments.

\subsection{Rehearsal}
\begin{figure}[]
      \includegraphics[width=0.9\linewidth]{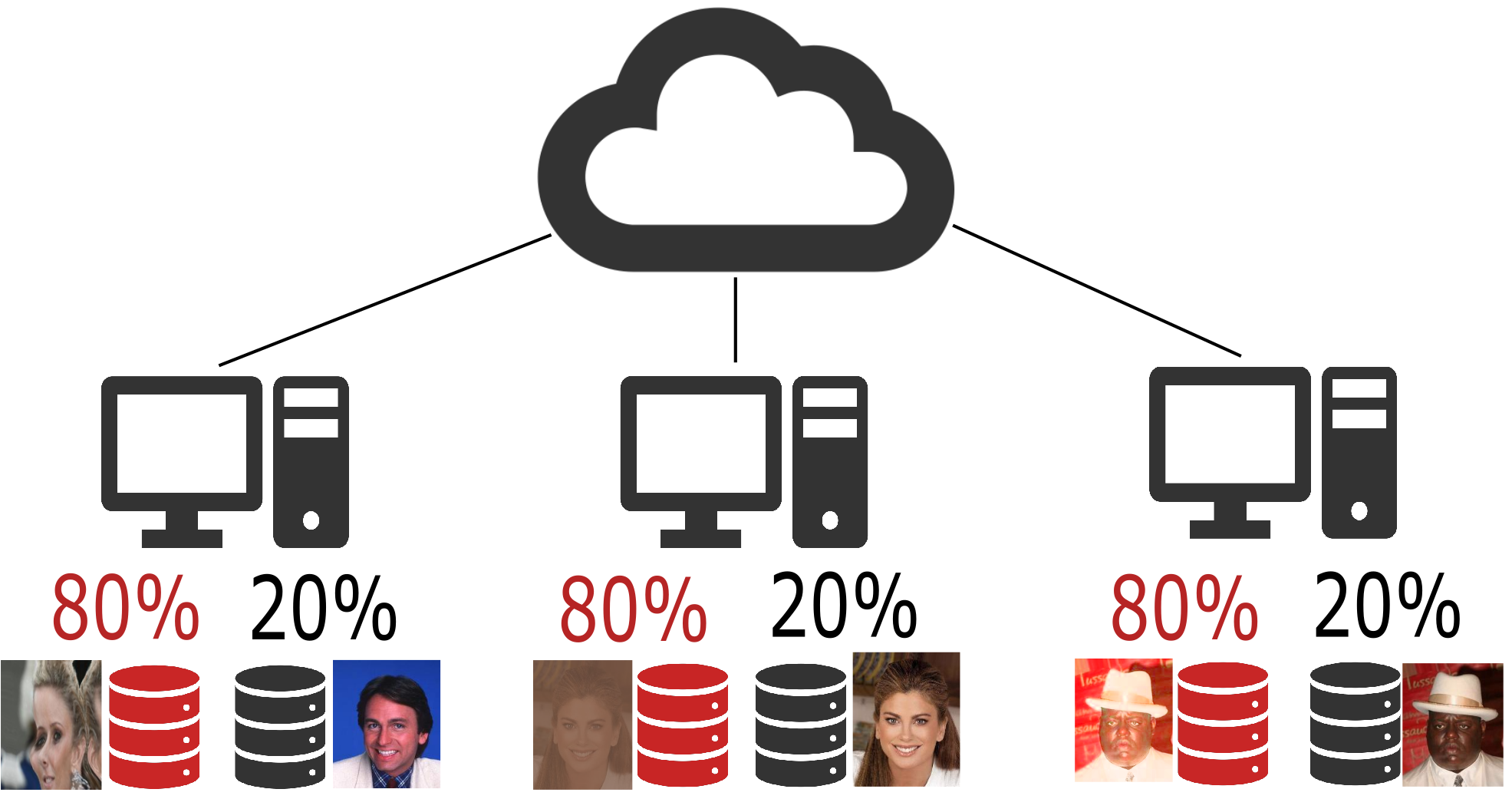}
      \caption{Rehearsal in FL: The clients mix their potentially shifted local train data with 20\% ID data for improvements in scenarios of CD and CF}
      \label{fig:rehearsal_method}
\end{figure}

\emph{Rehearsal}~\cite{rebuffi2017icarl} is a widely used method to handle CF in Continual Learning settings. It stores a small share of the data after each training stage and continuously interleaves old with new samples in future training epochs.
We apply this concept to a federated setting: \textbf{Each client trains on an additional certain share of data that remains untouched by the shifting transformation}. Using only local data, \textbf{ensures that no private data of a client is leaked to the global model or any other client}. The share of additional clean data makes up 20\% of the train data of each local client. 
Figure~\ref{fig:rehearsal_method} illustrates the concept of rehearsal in a federated setting.
We use our framework to test rehearsal in scenarios of CD as well as CF compared to the baseline with all clients shifted in CD and at the strongest shift strength level in the CD-scenario.

\begin{figure}[h]
      \includegraphics[width=1.0\linewidth]{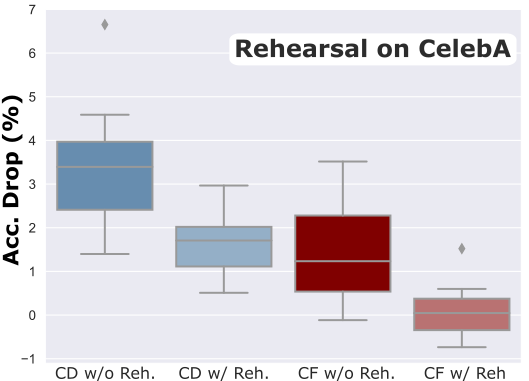}
      \caption{Performance impact from rehearsal on CD/CF}
      \label{fig:localrehearsal}
\end{figure}

Figure~\ref{fig:localrehearsal} shows the impact of rehearsal on the CD/CF Scenario. As expected from a method against CF, it decreases the performance drop on CelebA by \textbf{82.63\%} from 1.24\% to 0.12\%.
However, this commonly used CF-method also shows improvements in the CD Scenario, as it decreases the performance drop by \textbf{51.05\%} from 3.45\% on CelebA to only 1.69\%.
This motivates to jointly analyze both problems, as in Figure 9 of the main paper, which highlights the persistence of the \textbf{generalization bump}.

\subsection{Experiment Details}
\subsubsection{Parameter and Environment}
To improve reproducibility and make the setup more comprehensible, Table~\ref{tab:parameters} lists relevant parameters and details regarding the environment. It extends the written description from the main paper with additional details.

\subsubsection{Algorithms}
Algorithm~\ref{algo_cdcf} sketches the application of the joint CD/CF experiment as a pseudocode. $\mathit{max_{strength}}$ represents the maximum shift strength that the corresponding transformation can cause. In case of CelebA this means $\mathit{max_{strength}=5}$ with a step size of $1$ due to the fixed shift strength levels. For PESO, we assume $\mathit{max_{strength}}=1$ with an arbitrarily small step size due to its continuous scale.
To make the pseudocode more comprehensive, we don't include a possible efficiency improvement to the pseudocode: Skipping the initialization round for CD instead after the first CD-round would save many training epochs. However, this has no impact on the results and should be easier to understand.

\begin{algorithm*}[h]
\caption{Joint Analysis of CD and CF}
\label{algo_cdcf}
\textbf{Input: }$\mathit{seeds}$ 
\begin{algorithmic} 
\REQUIRE $\mathit{shifts_{cd}}=\left[0, 1\right]$ \\ \AND $\mathit{shifts_{cf}}=\left[0, \mathit{max\_strength}\right]$
\FOR{$\mathit{seed}$ in $\mathit{seeds}$}
    \FOR{$\mathit{shift_{cd}}$ in $\mathit{shifts_{cd}}$}
        \FOR{$\mathit{shift_{cf}}$ in $\mathit{shifts_{cf}}$}
            \STATE $\mathit{model} \leftarrow \mathit{init(seed)}$
            \STATE $\mathit{perf_{cdcf}} \leftarrow 0$
            \FOR{$\mathit{epoch}$ in $\mathit{cd\_epochs}$}
                \STATE $\mathit{model} \leftarrow train(\mathit{model}, \mathit{shift_{cd}}, \mathit{seed}, \mathit{shift\_strength}=1.0)$
            \ENDFOR
            \IF{$\mathit{shift_{cd}}==0$ \AND $\mathit{shift_{cf}}==0$}
                \STATE $\mathit{perf_{id}} \leftarrow metric(test(\mathit{model}, \mathit{data_{id}}))$
            \ENDIF
            \FOR{$\mathit{epoch}$ in $\mathit{cf\_epochs}$}
                \STATE $\mathit{model} \leftarrow train(\mathit{model}, \mathit{shift\_ratio=1.0}, \mathit{seed}, \mathit{shift_{strength}=shift_{cf}})$
            \ENDFOR
            \STATE $\mathit{perf_{cdcf}} \leftarrow metric(test(\mathit{model}, \mathit{data_{id}}))$
            \STATE $\mathit{\Delta perf_{cdcf}(shift_{cd}, shift_{cf})} \leftarrow \mathit{perf_{id}} - \mathit{perf_{cdcf}}$
        \ENDFOR
    \ENDFOR
\ENDFOR \\
\textbf{return }$\mathit{\Delta perf_{cdcf}}$

\end{algorithmic}
\end{algorithm*}

\begin{table*}[]
\begin{center}
\begin{tabular}{ |c|c|c| } 
 \hline
  & \textbf{CelebA} & \textbf{PESO} \\ 
  \hline
 Model & \emph{U-Net}~\cite{ronneberger2015u} & \emph{ViT}~\cite{vit} \\ 
 Model Implementation & \emph{ViT-FL~\cite{vitfl}}& \emph{BottleGAN}~\cite{bottlegan} \\
 Federated Algorithm & \emph{FedAvg}~\cite{fedavg} & \emph{FedAvg}~\cite{fedavg} \\
 Local Epochs & 1 & 1\\
 Global Epochs (CD) & 200 & 500 \\
 Global Epochs after shift (CF) & 100 & 40 \\
 Total Clients & 227 & 10 \\ 
 Clients per Global Epoch & 10 & 4\\
 Seeds & [1,2,3,4,5,6,7,8,9,10] & [1,2,3] \\
 Loss & Cross Entropy & Cross Entropy \\
 Optimizer & SGD & Adam \\
 Optimizer param. & - & $\beta_1=0.9$, $\beta_2=0.999$ \\
 Learning rate & 0.02 & 0.001 \\
 Step Size & 30 & 100 \\
 Batch Size & 32 & 48 \\
 Image/Patch Size & $178\times218$ & $114\times114$ \\
 GPU & NVIDIA GeForce RTX 4090 & NVIDIA A40 \\
 CUDA SDK Version & 12.2 & 11.6 \\
 PyTorch Version & 2.0.0 & 1.13.0 \\
 \hline
\end{tabular}
\caption{Parameters and Environment for the experiments on PESO}
\label{tab:parameters}
\end{center}
\end{table*}
\end{document}